\documentclass{article}

\usepackage[square, numbers]{natbib}
\usepackage[preprint]{neurips_2024}

\usepackage[utf8]{inputenc} 
\usepackage[T1]{fontenc}    
\usepackage{hyperref}       
\usepackage{url}            
\usepackage{booktabs}       
\usepackage{amsfonts}       
\usepackage{nicefrac}       
\usepackage{microtype}      
\usepackage{xcolor}         
\usepackage{graphicx}  
\usepackage{xcolor}        
\usepackage{makecell}
\usepackage{multicol}
\usepackage{multirow}
\usepackage{amsfonts}
\usepackage{algorithm}
\usepackage{algpseudocode}
\usepackage{amsmath}
\usepackage{amssymb}

\title{Learning Brain Tumor Representation in 3D High-Resolution MR
Images via Interpretable State Space Models}

\author{
  Qingqiao Hu\\
  Department of Computer Science\\
  Stony Brook University\\
  \texttt{qingqiao.hu@stonybrook.edu} \\
  \And
  Daoan Zhang \\
  Department of Computer Science\\
  University of Rochester\\
  \And
  Jiebo Luo \\
  Department of Computer Science\\
  University of Rochester \\
  \And 
  Zhenyu Gong\\
  Klinikum rechts der Isar\\
  Technical University of Munich \\
  \And
  Benedikt Wiestler \\
  School of Medicine and Health \\
  Technical University of Munich \\
  \And
  Jianguo Zhang \\
  Department of Computer Science\\
  Southern University of Science and Technology \\
  \And
  Hongwei Bran Li \\
  Martinos Center \\
  Harvard Medical School\\
  \texttt{holi2@mgh.harvard.edu} \\
  }

\begin{document}

\maketitle

\begin{abstract}
  Learning meaningful and interpretable representations from high-dimensional volumetric magnetic resonance (MR) images is essential for advancing personalized medicine. While Vision Transformers (ViTs) have shown promise in handling image data, their application to 3D multi-contrast MR images faces challenges due to computational complexity and interpretability. To address this, we propose a novel state-space-model (SSM)-based masked autoencoder which scales ViT-like models to handle high-resolution data effectively while also enhancing the interpretability of learned representations. We propose a latent-to-spatial mapping technique that enables direct visualization of how latent features correspond to specific regions in the input volumes in the context of SSM. We validate our method on two key neuro-oncology tasks: identification of isocitrate dehydrogenase mutation status and 1p/19q co-deletion classification, achieving state-of-the-art accuracy. Our results highlight the potential of SSM-based self-supervised learning to transform radiomics analysis by combining efficiency and interpretability. The code is available at \url{https://github.com/WinstonHuTiger/mamba_mae}.
\end{abstract}

\section{Introduction}
Learning effective and interpretable representations from high-dimensional volumetric magnetic resonance (MR) images is essential for advancing radiomics and personalized medicine in neuro-oncology \citep{scapicchio2021deep}. Particularly for brain tumors, capturing meaningful features from \emph{3D} and \emph{multi-channel} MR imaging is crucial for accurately characterizing their complex and heterogeneous nature \citep{yi2021current,lohmann2021radiomics}. These images' spatial complexity, high dimensionality, and variability highlight the need for advanced machine learning methods to extract not only discriminative but also interpretable radiomic features \citep{narang2016radiomics}.

Vision Transformers (ViTs) \citep{dosovitskiy2021image} have emerged as a promising solution for modeling global relationships in high-dimensional data through self-attention mechanisms \citep{liu2021swin}. However, applying ViTs to 3D MR images presents unique challenges. GPU memory limitations often necessitate resizing or cropping of volumetric images, resulting in significant information loss. Additionally, ViTs \citep{dosovitskiy2021image} struggle with processing long sequences, leading to the adoption of larger patch sizes, which can overlook fine-grained details which is essential for accurate radiomics analysis and understanding tumor characteristics.

To overcome these limitations, we propose a state-space model (SSM)-based architecture pre-trained using a masking strategy, designed for 3D high-resolution MR images. Our approach leverages the recent \emph{Mamba} architecture \citep{gu2023mamba,mamba2}, which excels at handling long sequences while maintaining linear computational complexity. This allows our model to scale effectively for high-resolution data, enabling it to capture both fine-grained and global features. Furthermore, to address the challenge of interpretability, we introduce a novel latent-to-spatial mapping technique that enables direct visualization of how different regions in the input volume contribute to the learned representations, providing clearer insights into the model's decision-making process.

We adopt a pretraining-finetuning paradigm to evaluate the performance of our SSM-based framework. Our experiments, which explore various input data sizes and patch dimensions, show that SSM-based models consistently outperform ViT-based models, particularly when processing larger input sizes. The superior performance of SSM models is attributed to their enhanced capacity for fine-grained sequence perception through efficient linear attention mechanisms. Notably, while ViTs face GPU memory constraints when processing detailed patches, our SSM-based models continue to improve in performance, achieving state-of-the-art results.

We validate our approach on two key neuro-oncology tasks: IDH mutation status and 1p/19q co-deletion classification. These genetic markers are crucial for glioma classification and patient stratification \citep{louis20212021}. Our results demonstrate the potential of SSM-based self-supervised learning representation, enabling efficient, interpretable, and accurate tumor characterization through high-dimensional imaging.

In summary, the contributions of this paper are threefold:
\begin{itemize}
    \item We proposed a novel SSM-based self-supervised learning framework for 3D multi-contrast MR images that linearly scales to high resolution.
    \item We introduce a latent-to-spatial mapping method that enhances the interpretability of learned representations by providing direct insights into how the model's features correspond to specific regions in the input volumes.
    \item We demonstrate state-of-the-art performance on IDH mutation and 1p/19q co-deletion classification tasks, showing the effectiveness of our approach in key neuro-oncology applications.
\end{itemize}

\section{Related Work}

\subsection{Image Representation for Glioma Classification}

Gliomas account for approximately 80\% of malignant central nervous system tumors \citep{ostorm2023neuro}. The 2021 WHO Classification guideline emphasizes the integration of genetic data, categorizing adult-type diffuse gliomas based on IDH mutation and 1p/19q co-deletion status \citep{louis20212021, horbinski2022clinical}. These genetic markers are crucial for prognosis and treatment planning \citep{tong2024idh1, Mohile2022therapy, Zhang2021}, yet their detection often relies on invasive biopsy methods \citep{ball2020frequency}.

Radiomics offers a promising non-invasive alternative, leveraging high-dimensional imaging data from routine MR scans to capture the complex biological behavior of gliomas \citep{singh2021radiomics, shui2021era}. However, the high dimensionality and intricate spatial structure of glioma radiomics pose significant computational challenges \citep{narang2016radiomics}. Our work addresses these challenges by introducing a novel SSM-based approach capable of efficiently processing high-resolution 3D MR images, improving the accuracy of non-invasive tumor subtyping.

\subsection{Self-Supervised Learning in Medical Imaging}

Self-supervised learning has emerged as a powerful tool for learning robust representation features, particularly valuable in medical imaging where labeled data can be scarce \citep{doersch2015unsupervised, wang2015unsupervised, noroozi2016unsupervised, zhang2016colorful, pathak2017learning}. Two primary strategies have been prominent:
(1) Contrastive Learning: This approach models image similarity and dissimilarity between multiple views of the data \citep{becker1992self, hadsell2006dimensionality, wu2018unsupervised, oord2018representation, chen2021exploring, grill2020bootstrap}. While effective, contrastive methods often rely heavily on data augmentation, which can be challenging in medical imaging due to the need for domain-specific transformations. 
(2) Masked Image Modeling: Pioneered by Masked Autoencoders (MAE) \citep{he2022masked}, MAE strategies task the model with predicting missing or masked portions of the input. 

Our work builds upon these self-supervised learning strategies, specifically adapting the MAE approach for 3D MR images. By incorporating state-space models, we extend the capabilities of existing methods to handle the long sequences inherent in high-resolution volumetric data.

\subsection{State Space Models}

State Space Models (SSMs) originated as probabilistic graphical models \citep{koller2009probabilistic}, which describe the relationship between latent states and observed measurements. Recently, SSMs have gained considerable attention for their capabilities in long-sequence modeling, especially within advanced neural networks \citep{s5smith2022simplified}.

The evolution of SSMs in deep learning began with HiPPO models \citep{hippogu2020}, which excelled at capturing long-range dependencies. However, early implementations such as LSSL encountered significant computational and memory limitations \citep{lsslgu2021}. This challenge was addressed with the Structured State Space (S4) model, which introduced normalized diagonal parameters, leading to more efficient and scalable SSM architectures \citep{dssgupta2022,s4dgu2022,s5smith2022simplified,liquids4hasani2022liquid,gssmehta2023long,megama2022mega,h3fu2022hungry}.

A major breakthrough came with the development of Mamba \citep{gu2023mamba}, also known as the Selectively Structured State Space Model (S6). Mamba employs a selective scanning mechanism, enabling efficient processing of input sequences while maintaining computational competitiveness with Transformer-based methods, particularly in natural language processing tasks. Mamba’s scanning algorithm has also been shown to be a specific case of general linear attention \citep{mamba2}, positioning it as an effective solution for vision tasks, including radiomics analysis. In addition to offering faster inference than Transformers, Mamba has achieved new benchmarks in a range of domains, including language, audio, and genomics \citep{gu2023mamba}.

In vision tasks, SSMs have been applied with notable success. For instance, Vision Mamba (Vim) reverses the input sequence and utilizes Mamba blocks in a manner similar to Vision Transformers (ViTs) \citep{zhu2024vision, dosovitskiy2021image}. VMamba improves on this by scanning input sequences at the patch level in multiple directions, enabling better detail capture \citep{liu2024vmamba}. Furthermore, Video Vision Mamba (Vivim) extends this approach to video sequences by introducing multiple scanning branches \citep{yang2024vivim}. ZigMa, on the other hand, introduces a plug-and-play module for multi-directional scanning in a DiT-style model \citep{hu2024zigmaa}.

Despite these advances, the application of SSMs in radiomics analysis remains largely unexplored. Our work addresses this gap by proposing a linearly scaled SSM-based model for radiomics, leveraging the Mamba architecture \citep{gu2023mamba}. This approach enables the efficient capture of both fine-grained and global features in 3D long-sequence modeling, a crucial capability for analyzing high-resolution medical imaging data. By harnessing SSMs' ability to manage long sequences with linear complexity, our novel framework for self-supervised learning advances the state-of-the-art in glioma classification and opens new possibilities for efficiently processing high-dimensional medical images, thereby offering transformative potential for radiomics analysis.

\section{Methodology}

\begin{figure*}
  \centering
  \includegraphics[width=0.85\textwidth]{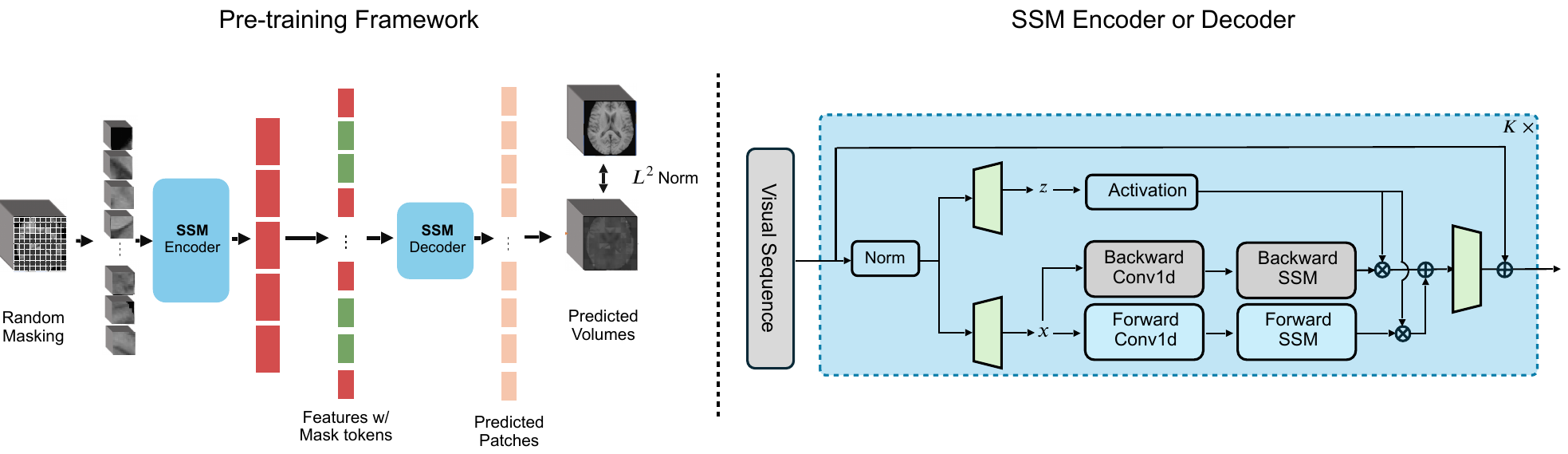}
\caption{\textbf{Left}: Pre-training a state space model to learn effective representations for 3D multi-contrast MR images. \textbf{Right}: Details of the SSM encoder and decoder. All the vanilla ViT's attention blocks in the masked autoencoder \citep{he2021masked} are replaced by SSM blocks, while preserving the same pre-training strategy by random masking and reconstruction. The scaled architecture effectively captures global representations for high-resolution data.}
  \label{fig:main_fig}
\end{figure*}

\subsection{Preliminaries on State Space Models}

State Space Models (SSMs) in deep learning, particularly the S4 model \citep{s4gu2021}, are designed to represent dynamical systems that evolve over time. These models utilize state variables that update based on their current values and the hidden states of the input, similar to how Recurrent Neural Networks (RNNs) process hidden states. However, SSMs offer a more efficient solution for capturing long-range dependencies, overcoming the limitations often associated with RNNs.

Formally, an SSM processes a one-dimensional input sequence $x(t) \in \mathbb{R}$ by transforming it into intermediate latent states $h(t) \in \mathbb{R}^N$. These latent states are subsequently used to generate the output $y(t) \in \mathbb{R}$. This transformation is governed by a discretized Ordinary Differential Equation:
\begin{align}
    h_t &= \bar{A}h_{t-1} + \bar{B}x_t, \\
    y_t &= Ch_t
\end{align}
Here, the matrices $\bar{A}$ and $\bar{B}$ are discretized using the zero-order hold method:
\begin{align}
    \bar{A} &= \exp(\Delta A), \\
    \bar{B} &= (\Delta A)^{-1}(\exp(\Delta A) - I) \cdot \Delta B
\end{align}
In this formulation, $A \in \mathbb{R}^{N \times N}$ represents the state matrix, while $B \in \mathbb{R}^{N \times 1}$ and $C \in \mathbb{R}^{N \times 1}$ are projection parameters. The timescale parameter is denoted by $\Delta$.

A significant difference between the S4 model and its advanced variant, the Selective State Space Model (S6) \citep{gu2023mamba}, is in how the parameters $B$, $C$, and $\Delta$ are handled. In S4, these parameters are not time-variant, while in S6, they dynamically adapt to the input at each time step, providing greater flexibility.

After discretization, SSMs are typically computed using global convolution:
\begin{align}
    \bar{K} &= (C\bar{B}, C\bar{A}B, \ldots, C\bar{A}^{k}B, \ldots), \\
    y &= x * \bar{K}
\end{align}
where $\bar{K} \in \mathbb{R}^L$ is a structured convolutional kernel, and $L$ represents the sequence length of the input. This convolutional approach differentiates S6 from RNN-like models, which rely on recurrent updates and are computationally more expensive when processing long sequences.

The \emph{Mamba} architecture, building blocks of S6, improves SSMs by incorporating selective state spaces. This mechanism enables \emph{Mamba} to focus on the most relevant parts of the input sequence, increasing efficiency in handling high-dimensional data. \emph{Mamba}'s selective processing of long sequences makes it particularly well-suited for tasks such as 3D MR image analysis in radiomics, where capturing both global and fine-grained details is crucial.

\subsection{3D Masked Image Modeling}

Following the ViTs' approach \citep{dosovitskiy2021image}, we divide 3D volumes into non-overlapping 3D patches. For instance, a $160 \times 160 \times 160$ volume can be segmented into 1000 patches, each of size $16 \times 16 \times 16$. This patch-based representation serves as the fundamental processing unit, effectively transforming 3D data into a 1D sequence for our SSM-based model.

We adopt a uniformly random masking strategy, inspired by \emph{MAE} \citep{he2021masked}. This approach leverages the inherent information redundancy in visual data and has demonstrated efficacy in representation learning, particularly when employing a high mask ratio. Unlike earlier methods such as inpainting \citep{pathak2016contex}, which focus on central region masking, our random approach mitigates center bias. This prevents the self-supervisory task from being trivially solved through extrapolation from visible neighboring patches, thereby encouraging the model to learn more robust and generalizable features.

\subsection{Linearly Scaled Model Architecture}

Our model's encoder, crucial for compressing 3D visual patches into latent representations, is based on the \emph{Vision Mamba} architecture \citep{zhu2024vision}. While the vanilla ViT \citep{dosovitskiy2021image} offers a global receptive field, it suffers from quadratic memory consumption in its self-attention layers \citep{dao2022flashattention}. This limitation becomes particularly problematic when processing large input volumes or using smaller patch sizes.

In contrast, our \emph{SSM}-based model \citep{gu2023mamba} achieves linear scaling, thanks to its global convolutional computation nature. To adapt the text-oriented vanilla S6 for visual sequences, we incorporate a backward branch in each S6 block, scanning the visual sequence bidirectionally (Fig.~\ref{fig:main_fig}). We extend this 2D design to a 3D version for volumetric data processing. Notably, we deviate from the vanilla ViT structure by replacing the class token with an average token calculation across the encoder, a strategy proven effective in recent vision models \citep{zhu2024vision,liu2024vmamba}.

The decoder reconstructs the input by utilizing both the encoded visible patches and masked tokens. Each mask token is a learnable vector, optimized to reveal the content of masked patches. Following insights from \citep{he2021masked} regarding the minimal importance of the decoder's embedding dimension, we employ a smaller decoder using the same architecture blocks as in the encoder. Both encoder and decoder incorporate absolute positional embeddings to maintain spatial information.

\begin{algorithm}
\caption{Latent-to-Spatial Mapping (PyTorch-like)}

\begin{algorithmic}[1]
\Require Tensor $Z \in \mathbb{R}^{N \times T \times E}$ 
\textcolor{blue}{\Comment{$N$: batch, $T$: tokens, $E$: embedding}}
\Ensure Tensor $\text{volume} \in \mathbb{R}^{N \times C \times H \times W \times D}$ 

\State $Z \gets Z.max(dim=-1)$ \textcolor{blue}{\Comment{Max pool on embedding}}
\State $Z \gets Z.unsqueeze(-1).repeat(1, 1, p^3)$ \textcolor{blue}{\Comment{Match patch size}}
\State $l, h, w \gets \text{round}(Z.shape[1]^{1/3})$
\State \textbf{assert} $l \times h \times w == Z.shape[1]$
\State $Z \gets Z.view(N, l, h, w, p, p, p, C)$ 
\State $Z \gets Z.permute(0, 7, 1, 2, 3, 4, 5, 6)$ 
\State $\text{volume} \gets Z.view(N, C, l \times p, h \times p, w \times p)$ 

\State \Return $\text{volume}$
\end{algorithmic}
\label{alg:unpatchify3d}
\end{algorithm}

\subsection{Interpretable Latent-to-Spatial Mapping}

While attention-based ViT models \citep{dosovitskiy2021image} offer straightforward interpretability of feature-space relationships, this correlation is less evident in SSM-based architectures. To address this limitation and enhance the interpretability of SSMs, we introduce a novel latent-to-spatial feature mapping technique by un-patchifying latent tokens to reconstruct the 3D volume, mapping features back to spatial locations. This method allows us to visualize and analyze how different regions in the input volume contribute to the model's latent representations.

Given 3D input volumes $X \in \mathbb{R}^{N \times C \times h \times w \times l}$, where $N$ is the batch size, $C$ is the number of input channels, and $h, w, l$ are the spatial dimensions, we define our model as $g(f(X))$. Here, $f(X)$ represents the encoder's SSM blocks, and $g(x)$ denotes either the decoder's SSM blocks or the classification head for downstream tasks.
Algorithm~\ref{alg:unpatchify3d} un-patchifies the latent feature $Z = f(X)$ and obtain the \emph{saliency volume} with same size of the input volume. 


Moreover, the \emph{Mamba} architecture, with its linear scaling and selective state-space mechanism, enables the use of smaller patch sizes. This capability allows the model to capture more detailed and localized information, enhancing the interpretability of the learned representations. 

\section{Experiments}

\subsection{Datasets}

We evaluate our linearly scaled SSM-based model using two public datasets. The first dataset, \textit{BraTS 2022} \citep{meze2015brats}, consists of 1251 multi-center MRI cases, providing a diverse foundation for initial model pre-training. The second dataset, the Erasmus Glioma Database (\textit{EGD}) \citep{egd2021}, includes 768 MRI scans and is used for fine-tuning and detailed evaluation. Both datasets include four modalities: FLAIR, T1, T2, and T1-contrast, with voxel dimensions standardized to $1 \times 1 \times 1$ mm. The \textit{BraTS} dataset is employed for pre-training, while the \textit{EGD} dataset is used for the downstream tasks of IDH mutation status (binary classification) and 1p/19q co-deletion (binary classification). Cases without requisite labels in the \textit{EGD} dataset are excluded from fine-tuning, ensuring data integrity. In total, 467 cases are available for IDH mutation classification, and 259 cases for 1p/19q co-deletion classification.

Pre-processing is standardized across datasets. A $160 \times 160 \times 160$ sub-volume centered on the brain’s center of mass is generated for each case. For volumes smaller than the target size, background padding is applied. This ensures uniformity in input dimensions, and an intensity normalization to the range $[0, 255]$ is performed to optimize the model for intensity-based augmentation techniques.

\subsection{Architecture and Training Configuration}

Our model is built upon the Vim blocks introduced by \cite{zhu2024vision}, integrated into both the encoder and decoder structures of the Masked Autoencoder (MAE) framework \citep{he2021masked}. The architecture is depicted in Figure~\ref{fig:main_fig}. The 3D input volumes are first divided into patches, and absolute 3D positional encodings are added to maintain spatial coherence, similar to the original MAE framework. The encoder contains 12 Mamba blocks, each with an embedding dimension of 384, while the decoder is streamlined with 8 Mamba blocks and an embedding dimension of 192.

The training process is divided into pre-training and fine-tuning stages. The pre-training phase runs for 1000 epochs with a learning rate of $1 \times 10^{-3}$ and weight decay of $0.05$. In the fine-tuning phase, only the encoder is updated over 100 epochs, with a reduced learning rate of $1 \times 10^{-4}$. A cosine learning rate scheduler is employed throughout both phases. Data augmentation techniques such as random affine transformations, Gaussian noise addition, and Gamma transformation are applied to improve generalizability.

\subsection{Evaluation Protocol}

The model’s performance is evaluated on the binary classification tasks using three key metrics: Accuracy (Acc), Area Under the Receiver Operating Characteristic Curve (AUC), and F1-score. A stratified 5-fold cross-validation protocol is employed, with 80\% of subjects in each class randomly allocated to the training set and the remaining 20\% used for testing. This ensures balanced evaluation across folds and provides a comprehensive assessment of the model’s predictive performance and generalizability.

\begin{table*}[t]
\centering
\small
\caption{5-fold cross-validation results for IDH classification and 1p/19q co-deletion classification using a linearly scaled model across different patch sizes $p \in \{4, 16, 32\}$ and comparison with the vanilla ViT and 3D ResNet pre-trained by MoCo-v3 \citep{chen2021empirical}}
\begin{tabular}{l c c c c c c c c}
\toprule
\multirow{2}{*}{\makecell[c]{Model \\ Name}}& \multirow{2}{*}{\makecell[c]{Patch \\ size} }& \multirow{2}{*}{\makecell[c]{Sequence \\ length} } & \multicolumn{3}{c}{IDH status classification} & \multicolumn{3}{c}{1p/19q co-deletion classification} \\
\cmidrule(r){4-6} \cmidrule(r){7-9}
& & & Accuracy & F1-score & AUC & Accuracy & F1-score & AUC \\
\midrule

Vanilla ViT &~16 & 1000 &0.931 & 0.894 & 0.967 & 0.860 & 0.713 & 0.897 \\
3D ResNet& - & - & 0.860 & 0.704 & 0.810 & 0.610 & 0.326 & 0.687 \\
\midrule 
\multirow{ 3}{*}{\makecell[c]{Linearly Scaled\\ Model}}&~32 & 125& 0.978  & 0.967  & 0.997  & 0.896  & 0.797  & 0.947 \\&~16 & 1000& 0.988 & 0.980 & 0.997 & 0.911 & 0.827& 0.944 \\ & ~4 & 64000 & \textbf{0.998} & \textbf{0.997} & \textbf{0.999} & \textbf{0.911} & \textbf{0.832} &  \textbf{0.958}\\

\bottomrule
\end{tabular}
\label{table:combined_classification}
\end{table*}

\section{Results}
\subsection{Linear Scaling Ability}

To evaluate the effectiveness of the linearly scaled model, we tested its performance with various patch sizes, $p$. Table~\ref{table:combined_classification} presents the results for two key classification tasks: IDH mutation status and 1p/19q co-deletion classification, using patch sizes of 4, 16, and 32. The performance of the linearly scaled model consistently improves as the patch size decreases, particularly for the more challenging 1p/19q co-deletion task \citep{egd2021}.

According to current literature, these results represent state-of-the-art performance: for instance, a multi-task U-Net \citep{van2022neuro} achieves an AUC of 0.900 for IDH mutation status and 0.850 for 1p/19q co-deletion, while a 2.5D multi-task U-Net \citep{chak2023neuro} achieves an AUC of 0.933 and 0.842, respectively. Additionally, we compared our method with a 3D ResNet featuring 18 residual convolutional blocks (shown in Table~\ref{table:combined_classification}) pre-trained by MoCo-v3 \citep{chen2021empirical}. The results indicate that the linearly scaled model, benefiting from its ability to capture both fine-grained and global features, possesses strong linear scaling capabilities.

We also compared the linearly scaled model with the vanilla ViT to validate the scalability of our design. The baseline vanilla MAE includes 12 attention layers with an embedding dimension of 384 and 12 attention heads in the encoder, alongside 8 layers with an embedding dimension of 192 and 16 attention heads in the decoder. Both models were trained under identical configurations. As shown in Table~\ref{table:combined_classification}, the linearly scaled model significantly outperforms the vanilla ViT in both classification tasks, attributed to the S6 model's superior ability to model long sequences. Notably, the vanilla MAE with patch sizes smaller than 16 could not be trained on 48GB Nvidia A8000 GPUs due to the quadratic memory and time complexity of self-attention relative to sequence length \citep{dao2022flashattention}.




\begin{table*}[t]
\centering
\caption{FLOPS and number of parameters of the encoder of both linearly scaled model and vanilla ViT with a 2D \(224 \times 224\) input image across different patch sizes $p \in \{4, 16\}$.}
\begin{tabular}{c cc c c}
\toprule
Backbone Name~& ~Patch size~&~Sequence length & FLOPS & Number of Parameters \\
\midrule
Mamba & 4 & 3136 & 534.98G & 13.75M \\
Mamba & 16 &196 & 34.46G & 12.89M \\ \hline
Transformer &4 & 3136 & 2520.17G & 22.52M \\
Transformer & 16& 196 & 73.73G & 21.67M \\
\bottomrule
\end{tabular}

\label{table:gflops_parameters}
\end{table*}

\subsection{Efficiency Analysis}

As previously mentioned, a 3D vanilla ViT with a patch size of 4 cannot fit into consumer-grade GPUs, requiring over 180GB of memory. To further evaluate the efficiency of the linearly scaled model, we assessed the floating-point operations per second (FLOPS) and the number of parameters of the encoder for both the linearly scaled model and vanilla ViT, using a 2D setting, as shown in Table~\ref{table:gflops_parameters}.

From Table~\ref{table:gflops_parameters}, it is evident that the SSM-based encoder scales roughly linearly ($\frac{534.96G}{34.4G} \approx \frac{3136}{196} = 16$) compared to the ViT-based encoder. This demonstrates that the SSM-based encoder efficiently utilizes GPU resources as sequence length increases, consistent with findings in previous works \citep{gu2023mamba,zhu2024vision}. The results highlight the superior scalability and resource efficiency of the linearly scaled model, particularly when processing longer sequences.

\begin{table*}
  \centering
  \caption{Robustness of linearly scaled model (patch size $p=16$) on IDH status classification.}
  \renewcommand{\arraystretch}{1.2} 
  \setlength{\tabcolsep}{4pt} 
  \scriptsize 
  \begin{tabular}{lccccccc}
  \toprule
  \textbf{Test-time Perturbation} & None & Rotation 10° & Rotation 45° & Rotation 90° & Bias Field (C=0.1) & Bias Field (C=0.4) & Bias Field (C=0.5) \\
  \midrule
  \textbf{Accuracy} & 0.990 & 0.979 & 0.968 & 0.968 & 0.990 & 0.990 & 0.625 \\
  \bottomrule
  \end{tabular}
  \label{table:robustness_egd}
\end{table*}

\begin{table*}[t]
\centering
\caption{Comparison of different pre-training methods, including no pre-training, contrastive learning (MoCo-v3 \citep{chen2021empirical}), and masked image modeling. Results are 5-fold cross-validation for IDH status classification and 1p/19q co-deletion classification.}
\begin{tabular}{l c c c c c c}
\toprule
\multirow{2}{*}{\makecell[c]{Pre-training method}} & \multicolumn{3}{c}{IDH status classification} & \multicolumn{3}{c}{1p/19q co-deletion classification} \\
\cmidrule(r){2-4} \cmidrule(r){5-7}
 & Accuracy & F1-score & AUC & Accuracy & F1-score & AUC \\
\midrule

No pre-training &0.943 & 0.911 &0.976 &0.725 & 0.202&0.589 \\
MoCo-v3 \citep{chen2021empirical}  & \textbf{0.995} & \textbf{0.994} & \textbf{0.999} & 0.888 & 0.791 & 0.914 \\
 Masked image modeling & 0.988 & 0.980 & 0.997 & \textbf{0.911} & \textbf{0.827}& \textbf{0.944} \\
\bottomrule
\end{tabular}
\label{table:pre_training_effectiveness}
\end{table*}

\subsection{Qualitative Results}

\begin{figure}
    \centering
    \includegraphics[width=0.48\textwidth]{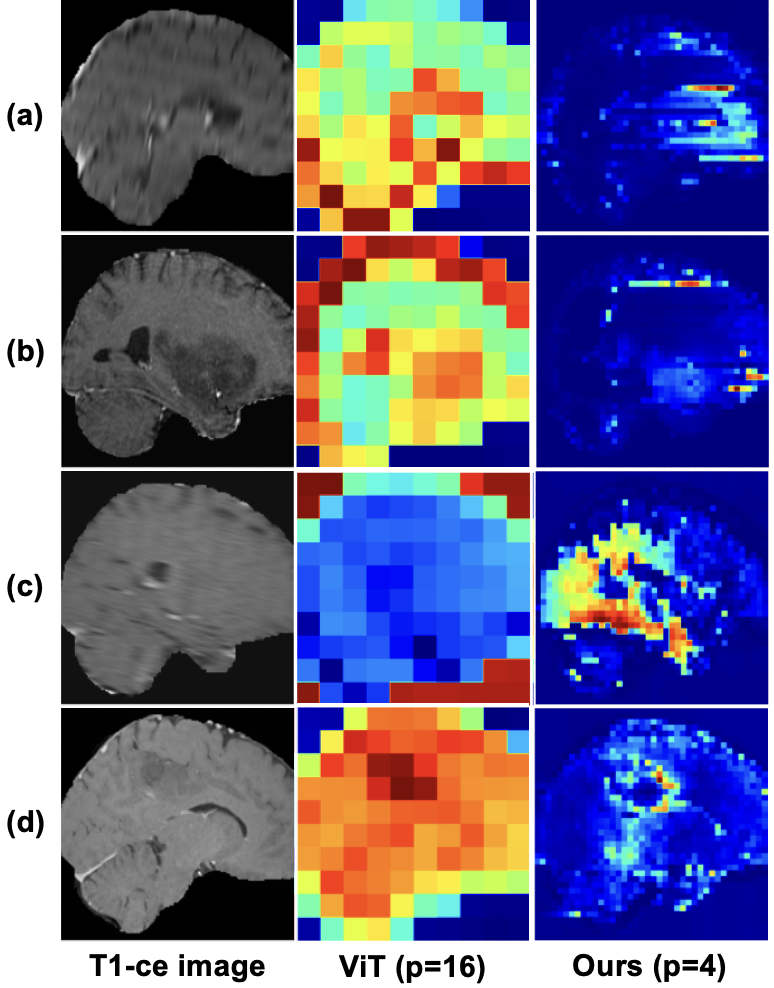}
   
    \caption{The comparison of saliency maps between the vanilla ViT with a patch size of 16 and our SSM-based model with a patch size of 4. (a) and (b) represent samples with IDH label 0 and 1, respectively; (c) and (b) represent samples with 1p/19q co-deletion label 0 and and 1, respectively}
    \label{fig:mamba_vis}
    
\end{figure}

We generated saliency maps from the fine-tuned versions of both our linearly scaled model (with a patch size of 4) and the vanilla ViT for comparison. To generate the saliency maps for the ViT, we followed the methodology described in \citep{oquab2024dinov}. As illustrated in Figure~\ref{fig:mamba_vis}, our model’s activations are predominantly concentrated around tumor regions, with the brighter areas indicating regions of interest. These areas highlight the tumor’s core and peripheral structures, which are crucial for the prediction of both 1p/19q co-deletion and IDH mutation status.

In contrast, the vanilla ViT model, using a larger patch size (16), exhibited more diffused activations, missing finer details that are essential for accurate classification. The smaller patch size processed by our model allowed it to focus on both global tumor structures and finer, local tissue details, providing a clearer and more interpretable representation. This capability to capture both global and fine-grained features explains the superior performance of our model, especially in tasks requiring high precision, such as 1p/19q co-deletion classification.

\subsection{Ablation Study}
\paragraph{Robustness Analysis.}
To evaluate the robustness of the linearly scaled model, we applied various perturbations, including rotations and bias field modifications \citep{SUDRE201750}, to the test-set data from one of the 5-fold cross-validation checkpoints. Table~\ref{table:robustness_egd} reports the impact of these perturbations on model accuracy.

While increasing the rotation degree results in a slight decrease in accuracy, the model remains generally robust. In contrast, applying a bias field transformation with a coefficient greater than 0.5 significantly degrades performance, with all predictions collapsing to zero. This highlights the model’s sensitivity to severe bias field distortions, emphasizing the need for careful data pre-processing and augmentation in real-world applications.

\paragraph{Effectiveness of Mask Image Modeling.} 
To demonstrate the effectiveness of the pre-training stage for the linearly scaled model, we compared its performance with different pre-training methods, including mask image modeling (ours), contrastive learning (MoCo-v3\citep{chen2021empirical}), and no pre-training. Table~\ref{table:pre_training_effectiveness} presents the classification performance across these methods for both IDH mutation status and 1p/19q co-deletion classification.

As shown in Table~\ref{table:pre_training_effectiveness}, pre-training significantly improves the performance of the model compared to no pre-training. While the MoCo-v3 method outperforms our method by a small margin on the IDH matutation status classification task (AUC of 0.995 compared to our 0.988), our mask image modeling method achieves more balanced performance across both tasks. This indicates that our method is not only effective but also more stable for a variety of radiomics analysis tasks, demonstrating its potential for broader deployment in medical imaging.

\section{Discussion}
Our work addresses critical challenges in handling high-resolution 3D MR imaging for radiomics analysis. By leveraging an SSM-based model, we achieve efficient processing of long sequences, allowing for superior feature extraction in glioma classification tasks.

Our work is not without limitations. A key component, the unpatchify algorithm, reconstructs the spatial structure from the latent space, yet it introduces challenges in preserving precise spatial correspondence. Due to information compression during patchification, finer spatial details critical for radiomics analysis may be lost. This compression can obscure subtle anatomical features, limiting the precision in mapping latent embeddings back to their original spatial locations. While our approach effectively captures global and fine-grained features, improving the spatial accuracy during reconstruction remains crucial. Future work should focus on refining the unpatchify process or developing alternative strategies to better preserve anatomical details, particularly for tasks requiring high spatial precision.

Additionally, improving the generalizability of the model, especially when exposed to different pre-processing variations and unseen data, remains an area for future exploration. Training the model on larger, more diverse medical imaging datasets could mitigate these issues and further enhance its applicability to downstream tasks such as segmentation and detection.

In conclusion, our model presents a promising approach for efficiently handling high-dimensional MR images in radiomics analysis. Future research will focus on refining spatial correspondence mechanisms, improving robustness to domain shifts, and continuing to optimize the model for large-scale clinical deployment.

\bibliography{egbib}
\end{document}